\documentclass[final]{cvpr}

\usepackage{times}
\usepackage{epsfig}
\usepackage{graphicx}
\usepackage{amsmath}
\usepackage{amssymb}

\usepackage{multirow}

\usepackage[pagebackref=true,breaklinks=true,colorlinks,bookmarks=false]{hyperref}

\begin{document}

\title{Adherent Mist and Raindrop Removal from a Single Image Using Attentive Convolutional Network}

\author{Da He\textsuperscript{1}, Xiaoyu Shang\textsuperscript{1}, and Jiajia Luo\textsuperscript{2}\thanks{Corresponding author.}\\
\textsuperscript{1}University of Michigan-Shanghai Jiao Tong University Joint Institute,\\Shanghai Jiao Tong University, Shanghai 200240, China\\
\textsuperscript{2}Biomedical Engineering Department, Peking University, Beijing 100191, China\\
{\tt\small \{da.he, shangxiaoyu\}@sjtu.edu.cn, jiajia.luo@pku.edu.cn}
}

\maketitle

\begin{abstract}
Temperature difference-induced mist adhered to the glass, such as windshield, camera lens, is often inhomogeneous and obscure, easily obstructing the vision and severely degrading the image. Together with adherent raindrops, they bring considerable challenges to various vision systems but without enough attention. Recent methods for other similar problems typically use hand-crafted priors to generate spatial attention maps.  In this work, we newly present a problem of image degradation caused by adherent mist and raindrops. An attentive convolutional network is adopted to visually remove the adherent mist and raindrop from a single image. A baseline architecture with general channel-wise attention, spatial attention, and multi-level feature fusion is used. Considering the variations and regional characteristics of adherent mist and raindrops, we apply interpolation-based pyramid-attention blocks to perceive spatial information at different scales. Experiments show that the proposed method can improve severely degraded images' visibility, both qualitatively and quantitatively. More applied experiments show that this underrated practical problem is critical to high-level vision scenes. Our method also achieves state-of-the-art performance on conventional dehazing and pure de-raindrop problems, in addition to our task of handling adherent mist and raindrops.
\end{abstract}

\section{Introduction}

\begin{figure}[t]
	\begin{center}
		\includegraphics[width=1\linewidth]{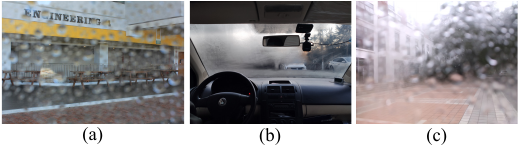}
	\end{center}
	\caption{Examples of adherent raindrops and/or mist. (a) Raindrops adhered to a glass before the camera \cite{1-qian2018attentive}; (b) Inhomogeneous mist adhered to a windshield when people sitting in the car; (c) Mist and raindrops simultaneously adhered to a camera lens. }
	\label{fig:1}
\end{figure}

High-level computer vision tasks, including object detection and segmentation, are highly dependent on the quality of captured images. However, outdoor vision systems such as security monitoring and vision-based self-driving cars or driving assistance systems are easily influenced by severe weather or other environments. As shown in Fig.~\ref{fig:1}(a), raindrops adhered to the camera lens or windshield can highly degrade the image or obstruct visibility. 

Adherent raindrops (or waterdrops) exist widely in different scenes. Without appropriate shelter, rain streaks will result in raindrops naturally. In environments like fishing-boats, splashing water can also form waterdrops. Raindrops have various visual effects because of various drop sizes, distances to the camera, and complex refraction, making it extremely difficult to model the raindrops manually. 

Besides raindrops, another severe and commonly-observed interference, adherent mist, is rarely mentioned in computer vision. As shown in Fig.~\ref{fig:1}(b), the mist adhered to the windshield can severely obstruct the vision and be dangerous for driving. The scene in Fig.~\ref{fig:1}(b) is easily encountered when driving a car in cold weather.

Interestingly, adherent mists sometimes form together with the raindrops, especially when driving a car with a temperature difference between the warm vapors inside the car and the glass cooled by the raindrops outside. Therefore, it is necessary to handle both adherent mist and raindrops at the same time. Fig.~\ref{fig:1}(c) shows the overlap of mist and raindrops adhered to a camera lens.

By improving the formulas in~\cite{1-qian2018attentive,2-chen2019gated,3-ren2018gated,4-li2019single}, the degraded image $I^{cap}$ captured by the camera can be mathematically described as a combination of raindrops $R$, the scattering of adherent mist $A$, and the clean background $B$ as follows:
\begin{align}
I^{cap} = ((1-M)\odot B + R)\odot t + A \odot (1-t)
\end{align}
where $M$ denotes to a binary mask showing the existence of raindrops, $\odot$ means element-wise multiplication, and $t$ is the transmission map indicating information passing rates through the adherent mist. $I^{cap}$, $M$, $B$, $R$, $t$, and $A$ are location-related maps (i.e., matrices) instead of constants.

Although disturbances of adherent mist and raindrops are widely observed in many scenarios and visibility can be critically deteriorated both for individuals and algorithms, very little work has been studied in the field of computer vision. Only a few articles have focused on the adherent raindrops recently~\cite{17-eigen2013restoring,1-qian2018attentive,19-quan2019deep}, while it seems that the adherent mist has not been mentioned. While both adherent mist and raindrops can be cleaned by hardware designs such as glass heaters and wipers, algorithm-based solutions need to be proposed for cost savings or automatic processing scenes.

In this study, we aim to remove the adherent mist and raindrops from a single impaired image to obtain a clean image. Based on convolutional neural networks (CNNs), the proposed approach can visually remove raindrops and mist of varying degrees. It can benefit high-level vision algorithms and improve their robustness in severe weather.

Since raindrops and mist may partially or even completely block some objects, it can be challenging to reproduce a purely clean image. To better solve the problem, we apply the basic blocks in~\cite{30-qin2019ffa} to construct our network. In addition to the typical convolutional layers, the basic block contains local residual learning, channel-wise attention, and spatial attention to adequately extract features and filter information. Long-distance residual shortcuts are also adopted in the CNN, thus dividing the basic blocks into several groups and then fusing their features. For strengthening the spatial attention and handling the variations of adherent mist and raindrops, an interpolation-based pyramid attention (IPA) block is applied to every group. Based on the same feature maps, the IPA block first zooms the feature maps to various sizes then perceives spatial attention maps from these scaled spaces.

In this study, we acquire an experimentally based dataset containing degraded and clean image pairs. Unlike purely synthetic datasets, the adherent mist and raindrops in our dataset exist and, therefore, can model the practical scenarios well. The proposed method can effectively improve the visibility of degraded images and outperforms other advanced networks. Moreover, it still exhibits state-of-the-art performance when our architecture is applied to traditional atmospheric haze removal or pure raindrop removal tasks.

In summary, our main contributions include:

	1) We address the visual interference problem caused by adherent mist and raindrops, which is frequent and essential but less studied. Without appropriate solutions, this problem limits extensive outdoor camera-based systems.
	
	2) An experimental-based dataset is acquired and can be utilized for further study, which contains 1560 image pairs, focusing on the coexistence of adherent mist and raindrops.
	
	3) An end-to-end restoration network is applied to effectively remove adherent mist and raindrops without hand-crafted priors. The IPA block contributes to handling images degraded by regional interferences (\eg, haze, adherent mist, raindrops). As a result, our network shows high restoration performance for the proposed task, the conventional de-hazing task, and the pure raindrop removal task.

\section{Related Work}
{\bf Rain Streak Removal. }Most visibility enhancement algorithms related to rain weather focused on rain streaks~\cite{4-li2019single,10-fu2017removing,11-li2017single,12-wang2019erl,13-zhang2018density}. However, the shapes and physical effects of raindrops exceedingly differ from those of rain streaks. Thus, these streak removal algorithms may not be suitable for raindrop removal problems.

\begin{figure*}[t]
	\begin{center}
		\includegraphics[width=1\linewidth]{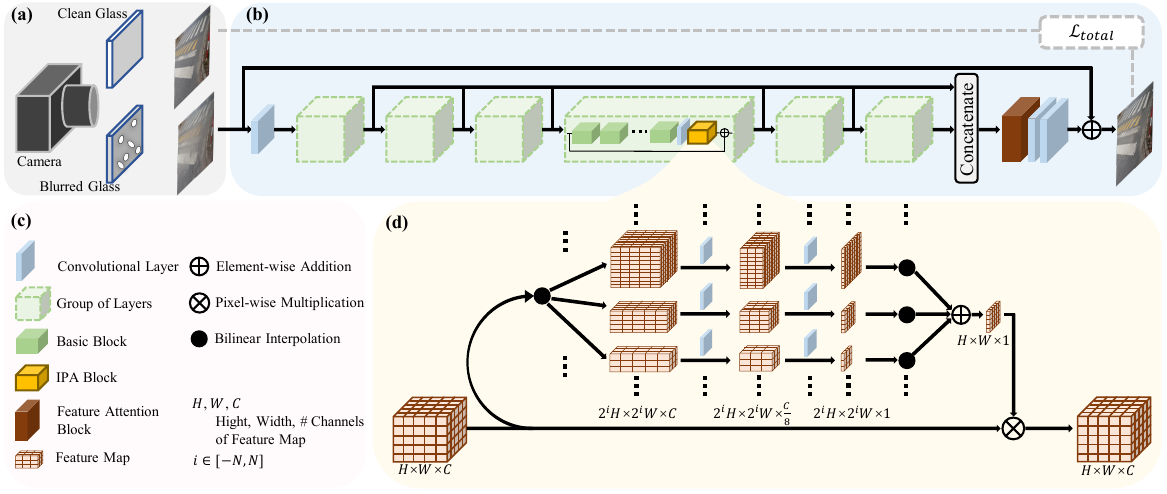}
	\end{center}
	\caption{Pipeline of the proposed method. (a) Diagram of the dataset acquisition; (b) Overview of the applied convolutional network; (c) Some legends; (d) The transformation of feature maps in the IPA block.}
	\label{fig:2}
\end{figure*}

{\bf Raindrop Removal. }A few studies about raindrop detection have been proposed for years. Kurihata \etal~\cite{14-kurihata2005rainy} utilize principal component analysis to learn the characteristics of raindrops, while Ito \etal~\cite{15-ito2015adherent} employ the maximally stable extremal regions to detect raindrop candidates. However, these methods do not concentrate on raindrop removal. You \etal~\cite{16-you2015adherent} pay attention to both raindrop detection and removal in videos. They notice that the temporal change of intensity of raindrop pixels is different from that of non-raindrop pixels, which does not apply to the single image case. Eigen \etal~\cite{17-eigen2013restoring} may be the first to address the single image raindrop removal problem. They use a straightforward idea of training a shallow CNN with pairs of raindrop images and clean images. Unfortunately, the limited capacity of the CNN model restricts its performance, especially for large and dense raindrops.

Qian \etal's work (DeRaindrop) in 2018~\cite{1-qian2018attentive} might be the first practical solution for the raindrop removal task. Specifically, they build a generative adversarial network (GAN)~\cite{18-goodfellow2014generative} to generate raindrop-free images. Besides the GAN, DeRaindrop includes a recurrent network to gradually locate raindrops’ location in the input image as a GAN's spatial attention. The performance of DeRaindrop is related to this raindrop mask, whose ground truth is calculated by the difference between the degraded image and the clean image with a hand-crafted threshold 30. However, it is difficult to apply this manual threshold to images with adherent mist, which contain complex pixel-value variations. Quan \etal~\cite{19-quan2019deep} propose shape-driven attention to the raindrop region based on priors about raindrops' shape.

{\bf De-hazing. }Few works primarily focus on the adherent mist. Some similar studies mainly concentrate on atmospheric haze (or fog)~\cite{2-chen2019gated,3-ren2018gated,20-cai2016dehazenet,21-fattal2008single,22-he2010single}. On the one hand, they both have a scattering effect, which results in a ``white” color and degrades the original information. However, compared to atmospheric haze, the adherent mist is more likely to be inhomogeneous as it is related to the specific design of the equipment, such as which direction the warm steam comes from and how the window looks (\eg, mist adhered to the windshield near the driver's seat in Fig.~\ref{fig:1}(b)).

Single image dehazing is initially explored using some priors-based methods. Fattal \etal~\cite{21-fattal2008single} work on estimating the albedo of the scene. He \etal~\cite{22-he2010single} propose a dark channel prior to discussing the local minimum of the dark channel varies between haze and haze-free images.

Then some learning-based methods have been proposed for dehazing using a single image. Cai \etal~\cite{20-cai2016dehazenet} introduce an end-to-end CNN to estimate the transmissions. Ren \etal~\cite{3-ren2018gated} propose a multi-scale gated fusion network to improve performance. And the smoothed dilation technique is adopted by Chen \etal~\cite{2-chen2019gated} to remove gridding artifacts. 

However, in our work, the superposition of adherent mist and raindrops results in complex local features. The influence of raindrops limits general de-hazing algorithms using transmission estimation. It is also difficult to manually input an attention map similar to that in~\cite{1-qian2018attentive} or~\cite{19-quan2019deep}. Therefore, the proposed network is designed to effectively utilize features and restore images for the practical scene.

\section{Method}
\subsection{Network Architecture}
The pipeline of our method is shown in Fig.~\ref{fig:2}. We follow~\cite{30-qin2019ffa} to build our network's main body, including its basic blocks and feature attention blocks. The main body, inspired by~\cite{30-qin2019ffa}, is denoted as a {\it baseline}. The {\it baseline} contains 114 basic blocks, uniformly distributed in 6 groups. There is a long-distance shortcut in each group to form the residual learning mechanism for every 19 basic blocks. The output feature maps of each group are then fused. A global residual shortcut is applied so that the entire network learns the transformation residuals.

To further improve the perception for adherent mist and raindrops, we insert an IPA block into each group in the {\it baseline} model to implement the final network (denoted as {\it baseline} + IPA)  as shown in Fig.~\ref{fig:2}(b).

\subsection{Interpolation-based Pyramid Attention}
Adherent mist and raindrops show patterns quite different from common human-made objects (\eg, buildings, chairs) and widely-studied noises (\eg, Gaussian noise, rain streaks). Human-made objects usually have relatively rigid shapes. The convolutional kernels are sensitive to their low-level edges and thereby high-level shapes. Conventional noises always cover relatively small areas, and most background information has remained. However, the adherent mist has non-rigid edges and effectively attenuates the sharpness of adherent raindrops in images. The adherent raindrops vice versa aggravate the visual inhomogeneity. Complex refraction and scattering happen to the coexistence of adherent mist and raindrops, causing much more severe visual degradation than the general atmospheric haze and pure raindrops. 

Therefore, additional techniques to help the CNN extract features from the non-rigid and severe degradation might be appreciated. Different from other pyramid attention mechanisms that fuse spatial attention for different layers~\cite{38-ni2020pyramid} or adopt different convolutional kernels for the same layer~\cite{36-li2018pyramid,37-huang2019mask,39-wang2019automated}, our IPA generates different scaling features by bilinear interpolations as shown in Fig.~\ref{fig:2}(d). Specifically, for the feature map from the same layer, IPA firstly interpolates the feature map, generating $N$ amplified feature maps and $N$ shrunk feature maps associate with the original feature map. Two convolutional layers are successively applied to each feature map, obtaining $2N+1$ attention maps with different scales. Finally, $2N$ attention maps are interpolated back to the original lateral size and added together with the attention map at the original size. The final spatial attention map is then adopted to weight the original feature map by pixel-wise multiplication.

Like other pyramid attention techniques, IPA learns spatial attention based on features with different scales, enabling larger receptive fields and more flexible perception for objects of various sizes. Besides, the interpolation operation outperforms convolution-based upsampling/downsampling (\eg,~\cite{35-mei2020pyramid,38-ni2020pyramid}) for another advantage in this task. Because the bilinear interpolation can be regarded as the low-pass filter when processing images, high-frequency information in the feature map is suppressed after interpolation. In the original feature map, the background textures in clean regions usually correspond to high-frequency information. In contrast, textures obscured by adherent mist and raindrops mainly contain relatively low-frequency information. Therefore, after interpolation (i.e., low-pass filter), the attention mechanism is less likely to pay meaningless attention to clean regions. Instead, areas covered by adherent mist and raindrops possibly obtain relatively more attention than before. Thus, the IPA block might be helpful to strengthen the attention mechanism for regional interference problems.

\subsection{Loss Function}
As indicated in~\cite{6-johnson2016perceptual,7-ledig2017photo}, although pixel-wise mean squared error (MSE) or mean absolute error (MAE) might bring significant improvements to metrics such as peak signal-to-noise (PSNR) and structural similarity index (SSIM)~\cite{9-wang2004image}, the generated image is likely to be over smoothed.  Instead, a perceptual loss term can be applied to calculate the error in high-level views, which might help generate subjectively realistic images. 

We calculate the perceptual loss based on the MSE of the output features from the third block of VGG16~\cite{7-ledig2017photo}. The VGG16 model is pretrained on the ImageNet dataset~\cite{29-deng2009imagenet}, which might give a relatively high-level description of an image. Finally, the perceptual loss is added together with the general MAE loss as:
\begin{align}
\mathcal{L}_{total} = \lambda_1\mathcal{L}_{mae} + \lambda_2\mathcal{L}_{per}
\end{align}
where$\lambda_1$, $\lambda_2$ are the weight coefficients of items, $\mathcal{L}_{total}$ is the total loss, $\mathcal{L}_{per}$ denotes the perceptual loss, and $\mathcal{L}_{mae}$ is calculated by directly comparing the prediction and the ground truth using MAE.

\subsection{Data Preparation}
To prepare the image pairs consisting of clean and degraded images, we take photos following strategies similar to those in~\cite{1-qian2018attentive}. The schematic is simplified in Fig.~\ref{fig:2}(a). Specifically, we utilize two same glass panels to simulate different cases. First, we use a tripod to secure the camera. Second, one of the panels is randomly sprinkled with waterdrops on one side and randomly sprayed with mist by a cosmetic sprayer on the other side. Then, we place the blurred panel in front of the camera and take a photo of the degraded image. Finally, the corresponding ground truth (i.e., clean) image is taken with the other clean glass panel.

The distributions of waterdrops and mist are random to keep good generalization. Moreover, the distance between the camera and the glass panel is also a random value from $0.5 cm$ to $4.0 cm$. Thus the shape of waterdrops can have different appearances in the image due to camera focal length. Canon EOS 800D and the integrated camera of Redmi K20 Pro are used for taking pictures.

Finally, all the collected images were resized with bicubic interpolation to the shape of $640\times480$ to form the dataset. There are 1248 image pairs for training, 156 pairs for validation, and 156 pairs for testing. All the following evaluation results are based on images in the testing set.

\begin{figure*}[t]
	\begin{center}
		\includegraphics[width=1\linewidth]{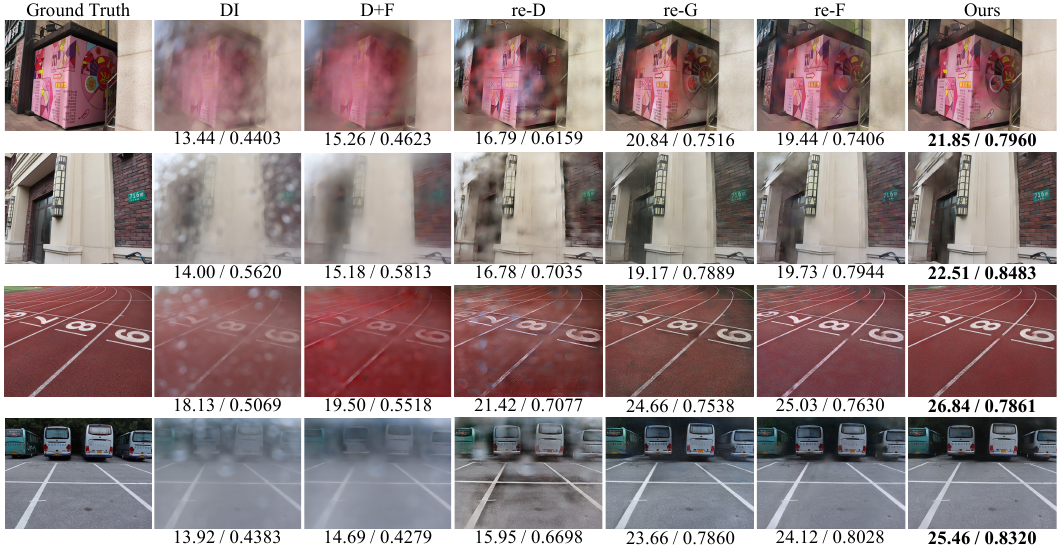}
	\end{center}
	\caption{Intuitive comparisons among a few restoration methods. From left to right: ground truth, degraded image (DI), D+F~\cite{1-qian2018attentive,30-qin2019ffa}, re-D~\cite{1-qian2018attentive}, re-G~\cite{2-chen2019gated}, re-F~\cite{30-qin2019ffa}, and our method. Each row corresponds to the same sample. The values below images (except ground truth images) indicate the corresponding values of PSNR / SSIM. The details of the images are best viewed by zooming in.}
	\label{fig:3}
\end{figure*}

\section{Experiment}
\subsection{Experiment Setup}
The proposed method is implemented using the PyTorch framework and trained with an Nvidia Titan RTX GPU. We use Adam optimizer with the learning rate of $1e-4$. The patch-based training strategy is applied with the cropped patch size of $240\times240$ and the batch size of 2. Other data augmentation operations including flipping and rotation are also adopted. In addition, for the IPA block, we set the parameter $N$ as $1$, and the coefficients $\lambda_1$, $\lambda_2$ in the loss function are $1.00$ and $0.04$, respectively.

\subsection{Restoration Results}
To our knowledge, there are no other comparable studies that work on the same problem (i.e., jointly handling adherent mist and raindrops). Thus, we compare the proposed method with the following three compromised aspects:

(1) Our method is compared with two two-step combinations. A two-step combination indicates we successively adopted a pre-trained dehazing approach (that originally focuses on atmospheric haze) and the pre-trained raindrop removal model DeRaindrop (D) by~\cite{1-qian2018attentive} to restore the image. In other words, the degraded image is firstly processed by a dehazing model (i.e., GCANet~\cite{2-chen2019gated} (G), or FFANet~\cite{30-qin2019ffa} (F)) and then by D. The two combinations are named as ``G+D'' and ``F+D'', individually.

(2) We also tried to apply D before G (or F) for comparison. These two inverse combinates are denoted as ``D+G'' and ``D+F'', individually.

(3) Although originally for other tasks, D, G, and F are advanced restoration networks that can be applied to our dataset. Therefore, we re-train these three networks using our dataset. The three re-trained methods are denoted as ``re-D'', ``re-G'', and ``re-F'', individually. For simplification, the degraded image in the test set is denoted as ``DI''. 

\begin{table}[t]
	\begin{center}
		\begin{tabular}{ccc}
			\hline
			Method & PSNR & SSIM \\
			\hline
			DI & 17.84 & 0.6048 \\
			\hline
			G+D & 15.12 & 0.5919 \\
			F+D & 17.57 & 0.6302 \\
			\hline
			D+G & 16.30 & 0.6167 \\
			D+F & 18.83 & 0.6383 \\
			\hline
			re-D & 18.89 & 0.7012 \\
			re-G & 22.88 & 0.7968 \\
			re-F & 23.45 & 0.8056 \\
			\hline
			Ours & {\bf 24.66} & {\bf 0.8293} \\
			\hline
		\end{tabular}
	\end{center}
	\caption{Quantitative comparisons of different methods on our dataset for adherent mist and raindrops removal tasks.}
	\label{table:1}
\end{table}

The results of Table~\ref{table:1} show that, firstly, it is important to handle adherent mist and raindrops together. The co-existence of mist and raindrops might lead to terrible results that are even worse than the degraded images (\eg, G+D results in Table~\ref{table:1}). Besides, the three re-trained models generally show effectiveness, especially for re-G and re-F, but our method shows the best performance.

In realistic scenes, the combination of adherent mist and raindrops might severely impair the image, blocking many objects from the subjective view. Examples in Fig.~\ref{fig:3} exhibit the awful shooting environments and the restoration results from several methods above. Specifically, due to space limitations, we only present five relatively high-performance methods (i.e., D+F, re-D, re-G, re-F, Ours) in Fig.~\ref{fig:3} as the intuitively compared methods.

According to Fig.~\ref{fig:3}, we can find that some objects are not distinguishable in the degraded image. For example, in the first row of Fig.~\ref{fig:3}, the decorative text on the pink wall is invisible in the degraded image. By contrast, most words can be distinguished in the restoration result from our method. A similar phenomenon is shown in the second row. The key information of the captured image, the house number, is amazingly recovered by our method. For the third sample, our result's general color is intuitively closest to that in the ground truth. The example in the fourth row can be regarded as a traffic control application. Suppose the camera is severely influenced by the terrible environment and takes degraded photos as shown in the picture. In this case, high-level algorithms may not recognize the plate number and guards cannot easily confirm the bus position. However, our method can recover the numbers on those bus bodies and observe the condition of the parking lot at a glance.

As for comparison, the proposed method can obtain relatively cleaner images and fewer artifacts than the other methods, which is also proved according to the attached PSNR / SSIM values. Besides, the combination method (i.e., D+F) in Fig.~\ref{fig:3} shows tiny improvement. In addition, our method is also robust for various conditions. Both relatively weak interference (\eg, the third sample in Fig.~\ref{fig:3}) and relatively severe degradation (\eg, the other three samples in Fig.~\ref{fig:3}) can be handled. The fact that the color of the cement floor in the fourth sample in Fig.~\ref{fig:3} shows similar color to the mist does not impair our method.

Therefore, the severe degradation of visibility from adherent mist and raindrops is presented. Unlike the well-studied restoration problems like de-noise or motion blur removal, the new task above is much more complicated and it is hard to estimate the degradation kernel. Simply re-training restoration networks remains many artifacts. Our method outperforms the compared advanced networks, quantitatively and intuitively. 

\subsection{Extended Experiments}
Besides the newly proposed vision task  described above, we also extend our network to other two conventional problems to show the broad applicability. Specifically, the proposed network is re-trained for atmospheric de-hazing and pure adherent raindrop removal applications.

\begin{figure}[t]
	\begin{center}
		\includegraphics[width=1\linewidth]{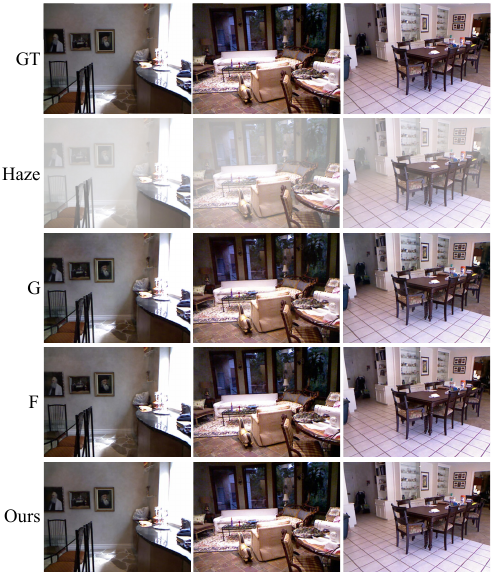}
	\end{center}
	\caption{Qualitative comparisons for a few de-hazing methods. From top to bottom: ground truth (GT), hazy image, G~\cite{2-chen2019gated}, F~\cite{30-qin2019ffa}, and our method. Each column corresponds to the same sample. }
	\label{fig:4}
\end{figure}

{\bf De-hazing. }Without any modification to the architecture, we also train our network using the RESIDE dataset, a widely-adopted dehazing benchmark. Following the same operations as previous studies, we adopt the indoor set containing over 13K image pairs for training and the indoor synthetic objective testing set (SOTS) including 500 pairs for testing. The testing results are listed in Table~\ref{table:2}. For fairness and convenience, the performance values of other methods are cited from~\cite{30-qin2019ffa}. In addition, three samples from the testing set are applied in Fig.~\ref{fig:4} for qualitative comparison. Due to space limitations, only our method and the two relatively new advanced approaches (i.e., G~\cite{2-chen2019gated} and F~\cite{30-qin2019ffa}) are presented.

\begin{table}[t]
	\begin{center}
		\begin{tabular}{ccc}
			\hline
			Method & PSNR & SSIM \\
			\hline
			DCP~\cite{22-he2010single} & 16.62 & 0.8179 \\
			AOD-Net~\cite{33-li2017aod} & 19.06 & 0.8504 \\
			DehazeNet~\cite{20-cai2016dehazenet} & 21.14 & 0.8472 \\
			GFN~\cite{3-ren2018gated} & 22.30 & 0.8800 \\
			G~\cite{2-chen2019gated} & 30.23 & 0.9800 \\
			F~\cite{30-qin2019ffa} & 36.39 & 0.9886 \\
			\hline
			Ours & {\bf 40.02} & {\bf 0.9931} \\
			\hline
		\end{tabular}
	\end{center}
	\caption{Quantitative comparisons of different methods on SOTS for de-hazing tasks.}
	\label{table:2}
\end{table}

As shown in Fig.~\ref{fig:4}, both the previous works and our method achieve nearly perfect performance. However, according to Table~\ref{table:2}, the proposed method outperforms other methods for PSNR and SSIM metrics. Therefore, the proposed method could be applied to handle the conventional de-hazing problem with appropriate training.

\begin{figure}[t]
	\begin{center}
		\includegraphics[width=1\linewidth]{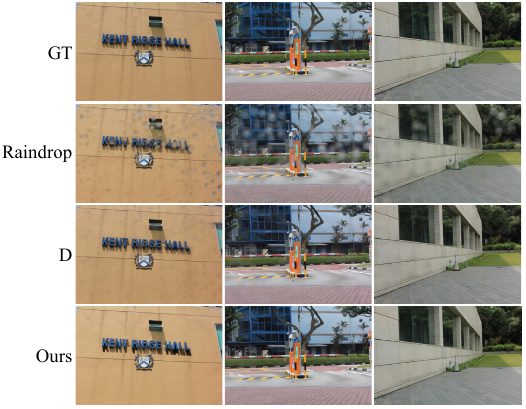}
	\end{center}
	\caption{Qualitative comparisons for two raindrop removal methods. From top to bottom: ground truth (GT), image with raindrops, D~\cite{1-qian2018attentive}, and our method. Each column corresponds to the same sample. Some fine artifacts are best compared by zooming in. }
	\label{fig:5}
\end{figure}

{\bf Pure Raindrop Removal. }As mentioned in Section 2, DeRaindrop (D)~\cite{1-qian2018attentive} is a milestone and a state-of-the-art approach for handling adherent raindrops. A dataset with pure raindrop degraded image and clean image pairs was also published by~\cite{1-qian2018attentive}. Therefore, we also try our method using that dataset to verify our approach's feasibility when only handling adherent raindrops. The dataset contains 861 image pairs for training and 58 image pairs for quantitative evaluation. We still do not modify the architecture of our method. The quantitative performance is listed in Table~\ref{table:3}. Similarly, the metric values of compared methods are directly cited from~\cite{1-qian2018attentive}. Three examples from the testing set are used for qualitative comparison in Fig.~\ref{fig:5}.

According to Table~\ref{table:3}, our method achieves comparable results with D. Although our PSNR value is slightly lower than that of D, our method shows the highest SSIM value.

As for the quantitative performance, our method results in more harmonious colors than those from D, which can be observed in Fig.~\ref{fig:5}. By carefully comparing the walls in the restoration images of D, it is not difficult to find that some light ``shades'' remain, which correspond to the original raindrops' locations. By contrast, our method does much better for this issue.

\begin{table}[t]
	\begin{center}
		\begin{tabular}{ccc}
			\hline
			Method & PSNR & SSIM \\
			\hline
			Eigen13~\cite{17-eigen2013restoring} & 28.59 & 0.6726 \\
			Pix2Pix~\cite{34-isola2017image} & 30.14 & 0.8299 \\
			D~\cite{1-qian2018attentive} & {\bf 31.57} & 0.9023 \\
			\hline
			Ours & 31.33 & {\bf 0.9297} \\
			\hline
		\end{tabular}
	\end{center}
	\caption{Quantitative comparisons of different methods on the test set in~\cite{1-qian2018attentive} for pure adherent raindrop removal tasks.}
	\label{table:3}
\end{table}

\begin{figure}[t]
	\begin{center}
		\includegraphics[width=1\linewidth]{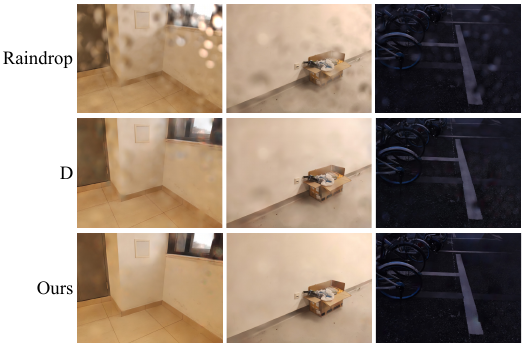}
	\end{center}
	\caption{Qualitative comparisons for two raindrop removal methods under specific illumination conditions. From top to bottom: image with raindrops, D~\cite{1-qian2018attentive}, our method. Each column corresponds to the same sample. Some artifacts are best compared by zooming in. }
	\label{fig:6}
\end{figure}

\begin{figure*}[t]
	\begin{center}
		\includegraphics[width=1\linewidth]{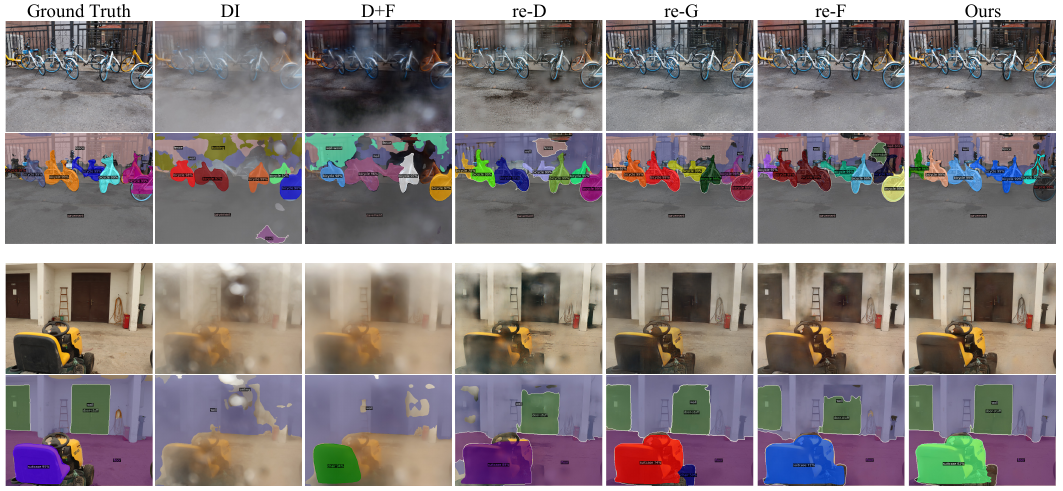}
	\end{center}
	\caption{Restoration results from different methods and their panoptic segmentation results. From left to right: ground truth (GT), degraded images (DI), D+F~\cite{1-qian2018attentive,30-qin2019ffa}, re-D~\cite{1-qian2018attentive}, re-G~\cite{2-chen2019gated}, re-F~\cite{30-qin2019ffa}, and our method. Each two rows correspond to the same sample. The panoptic segmentation result is attached below the corresponding image. The details of the images are best viewed by zooming in.}
	\label{fig:7}
\end{figure*}

The analysis of the above issue is followed. Our method might be more practical for various scenes because DeRaindrop set a threshold of 30 to segment raindrop mask for training, which is hand-crafted prior and might not be suitable to scenes with relatively monochromatic backgrounds. In these scenes, the pixel difference between raindrop regions and clean regions might become small, thereby the prior does not work and the performance of D decreases. By contrast, no hand-crafted thresholds are applied during our training. Therefore the proposed method can effectively learn features by itself with the dataset containing all kinds of illumination or background conditions. The above analysis might be qualitatively proved in Fig.~\ref{fig:6}. The three degraded images in Fig.~\ref{fig:6} are newly captured by our smartphone and do not exist in the dataset proposed by~\cite{1-qian2018attentive}. Both models of D and our method are only trained on the training set of the dataset in~\cite{1-qian2018attentive} (i.e., the same as before in Table~\ref{table:3}). Images in the first two columns mainly show a warm color tone and the last sample seems dark due to low illumination. Our restoration results show fewer artifacts than those from D.

{\bf Summary of the Extended Experiments. }Although the primary purpose of this article is to call for attention on the newly proposed problem: adherent mist and raindrop removal, the proposed method works well not only on our own dataset containing adherent mist, but also for conventional de-hazing and pure raindrop removal problems. In other words, these several tasks can be effectively handled in applications using only one method.

\subsection{Ablation Study}
Our network's architecture can be divided into two parts: the {\it baseline} network and the IPA blocks, as mentioned in Section 3.1. Therefore, we investigate the influence of the existence of IPA blocks concerning the above problems. The results of the ablation study can be found in Table~\ref{table:4}.

\begin{table}[h]
	\begin{center}
		\begin{tabular}{cccc}
			\hline
			Task & Metrics & {\it baseline} & {\it baseline}+IPA \\
			\hline
			\multirow{2}{*}{Proposed Task} & PSNR & 23.45 & {\bf 24.66}\\
					& SSIM & 0.8056 & {\bf 0.8293} \\
			\hline
			\multirow{2}{*}{De-hazing} & PSNR & 36.54 & {\bf 40.02} \\
					&SSIM & 0.9895 & {\bf 0.9931} \\
			\hline
			\multirow{2}{*}{Raindrop Removal} & PSNR & 30.82 & {\bf 31.33} \\
					& SSIM & 0.9247 & {\bf 0.9297} \\
			\hline
		\end{tabular}
	\end{center}
	\caption{Quantitative results of the ablation study for three tasks.}
	\label{table:4}
\end{table}

Therefore, Table~\ref{table:4} quantitatively proves that IPA blocks contribute to performance improvement for all the three experiments described before.

\subsection{Application Experiment}
Besides the above quantitative and qualitative evaluation results, the importance of the proposed problem and the effectiveness of our method are also validated in practical applications. Panoptic segmentation is a newly proposed segmentation task~\cite{32-kirillov2019panoptic} that tries to label all the image pixels. It is an essential high-level task for various industries like self-driving. The segmentation results based on ground truth images, degraded images, and restoration images are shown in Fig.~\ref{fig:7}. Specifically, we applied the panoptic segmentation algorithm, R101-FPN~\cite{31-cai2019cascade} from Detectron2 (https://github.com/facebookresearch/detectron2) model zoo to segment these images.

As shown in Fig.~\ref{fig:7}, panoptic segmentation might be easy to show the influence of adherent mist and raindrops from the view of computer vision algorithms. If observing the first degraded image, the segmentation algorithm cannot find all bicycles and is not sure what is behind bicycles. Undoubtedly the uncertainty may influence the decision of self-driving cars. Fortunately, bicycles could be counted correctly from images obtained by the last four methods and the fence could be roughly located in our restoration result. The second sample shows a semi-indoor environment. The segmentation borders are not clear in the degraded image. Simultaneously, the last four methods could restore the image to the extent that the high-level algorithm can approximately find the edges of different surfaces.

Therefore, the results of the application experiment show that mist and raindrops adherent to camera lenses or windshields can severely hamper high-level computer applications. The proposed algorithm-based solution might help ease the problem and make the visual devices more robust to some extent.

\section{Conclusion}
In this work, we might be the first to propose the image degradation problem related to the co-existence of adherent mist and raindrops. Both interference sources are attached to the lens or glass window of cameras, showing different image characteristics compared with atmospheric haze or rain streaks. Although the problem is widely observed in reality, little works have been investigated. We adopt an attentive convolutional network containing interpolation-based pyramid attention blocks to strengthen the spatial perception of regional obstacles to handle this problem. Both quantitative evaluation and intuitive examples show quality improvement with the proposed method without explicit priors, even for conventional de-hazing and pure raindrop removal tasks. The restoration from the adherent mist and raindrop degradation can benefit high-level tasks such as panoptic segmentation for producing reliable results, which demonstrates the vital significance in vision applications.

{\small
\bibliographystyle{ieee_fullname}
\bibliography{egbib}
}

\end{document}